\documentclass{article}
\pdfoutput=1

\usepackage[numbers]{natbib}
\usepackage[preprint]{neurips_2021}
\usepackage{array}
\usepackage{nicefrac}

\usepackage[utf8]{inputenc} 
\usepackage[T1]{fontenc}    
\usepackage{hyperref}       
\usepackage{url}            
\usepackage{booktabs}       
\usepackage{amsfonts}       
\usepackage{nicefrac}       
\usepackage{microtype}      
\usepackage{xcolor}         
\usepackage{subfloat}

\usepackage{todonotes}                                                           
\usepackage{comment}
\usepackage{multirow}
\usepackage[title]{appendix}
\usepackage[ruled,vlined]{algorithm2e}
\usepackage{algorithmic}
\usepackage{mathtools}
\usepackage{floatrow}
\usepackage{wrapfig}

\newcommand\vmaze{\texttt{Visual Maze}}
\newcommand\vsud{\texttt{Visual Sudoku}}

\newcommand\NSN{ \texttt{NSNnet}}

\newcommand\SATNET{\texttt{SAT-Net}}
\newcommand\VIN{\texttt{VIN}}

\newcommand\reinf{\texttt{REINFORCE}}
\renewcommand{\mathcal}[1]{#1}
\newcommand\numberthis{\addtocounter{equation}{1}\tag{\theequation}}

\title{End-to-End Neuro-Symbolic Architecture for Image-to-Image Reasoning Tasks}

\author{%
  Ananye Agarwal \\
  IIT Delhi\\
  \texttt{ananayagarwal@gmail.com} 
  \And
    Pradeep Shenoy \\
  Google Research India\\
  \texttt{shenoypradeep@google.com} 
  \And
    Mausam \\
  IIT Delhi\\
  \texttt{mausam@cse.iitd.ac.in} 
}

\begin{document}

\maketitle

\begin{abstract}
  Neural models and symbolic algorithms have recently been combined for tasks requiring both perception and reasoning. Neural models ground perceptual input into a conceptual vocabulary, on which a classical reasoning algorithm is applied to generate output. A key limitation is that such neural-to-symbolic models can only be trained end-to-end for tasks where the output space is symbolic. In this paper, we study \emph{neural-symbolic-neural} models for reasoning tasks that require a conversion from an image input (e.g., a partially filled sudoku) to an image output (e.g., the image of the completed sudoku). While designing such a three-step hybrid architecture may be straightforward, the key technical challenge is \emph{end-to-end} training – how to backpropagate without intermediate supervision through the symbolic component. We propose \NSN, an architecture that combines an image reconstruction loss with a novel output encoder to generate a supervisory signal, develops update algorithms that leverage policy gradient methods for supervision, and optimizes loss using a novel subsampling heuristic. We experiment on problem settings where symbolic algorithms are easily specified: a visual maze solving task and a visual Sudoku solver where the supervision is in image form. Experiments show high accuracy with significantly less data compared to purely neural approaches. 
\end{abstract}

\section{Introduction}

The human brain has long served as functional and architectural inspiration for the design of machine learning systems, from highly stylized abstractions of biological neurons~\citep{Rosenblatt1958,Rumelhart1986} to proposals of dichotomous systems of intelligence (``fast/instinctive  and slow/deliberative'' \citep{Kahneman2011}). Artificial neural networks show universal approximability, handle uncertainty, and support end-to-end training, making them the model of choice for many tasks processing noisy, ambiguous real-world inputs such as language and vision. Unfortunately, they do not yet match up to traditional algorithms on cognitive reasoning tasks such as Sudoku. Current neural reasoning  \citep{Tamar2016,Wang2019,Payani2019} models are data-intensive, approximate, and generalize poorly across problem sizes.

A recent line of inquiry combines neural and symbolic approaches for perceptuo-reasoning tasks~\cite{Tsamoura2021,Manhaeve2018, Yang2020}. This approach combines the core strengths of each individual system -- neural systems ground noisy, uncertain, complex perceptual signals into abstract symbols, on which symbolic reasoners perform exact, algorithmic computations to produce the desired (symbolic) output. Such neural-to-symbolic (NS) models offer many advantages over purely neural systems; in particular they are interpretable, and can be trained with limited data in an end-to-end fashion (i.e., with no supervision on the intermediate abstract problem constructed from the perceptual input); however, they are only applicable for problems where the output space is symbolic.

Our goal is to push the boundaries of such neuro-symbolic systems, by studying reasoning tasks where both inputs and outputs are in the physical/perceptual layer. We believe this is a small but necessary step for broadening the mechanisms by which embodied agents can interact with and learn from the natural world. We study image to image reasoning tasks such as \vsud, where partially- and fully filled Sudoku boards are presented as images, and \vmaze\ where the output image overlays the shortest path on the maze. We naturally extend NS models to \emph{neural-symbolic-neural} (NSN) architectures, where the symbolic solution is converted by an additional neural layer to produce an image. A key technical challenge in realizing such an architecture is end-to-end (E2E) training. E2E training regimes are useful for reducing the need for data annotation and are a particular strength of modern ML systems.
But, since there is a symbolic component in the \emph{middle} of the model, it is not clear how to backpropagate the loss from the output image to the input encoder.

\begin{figure}[t]
\label{fig:formulation}
    \centerline{
        \includegraphics[width=0.35\linewidth]{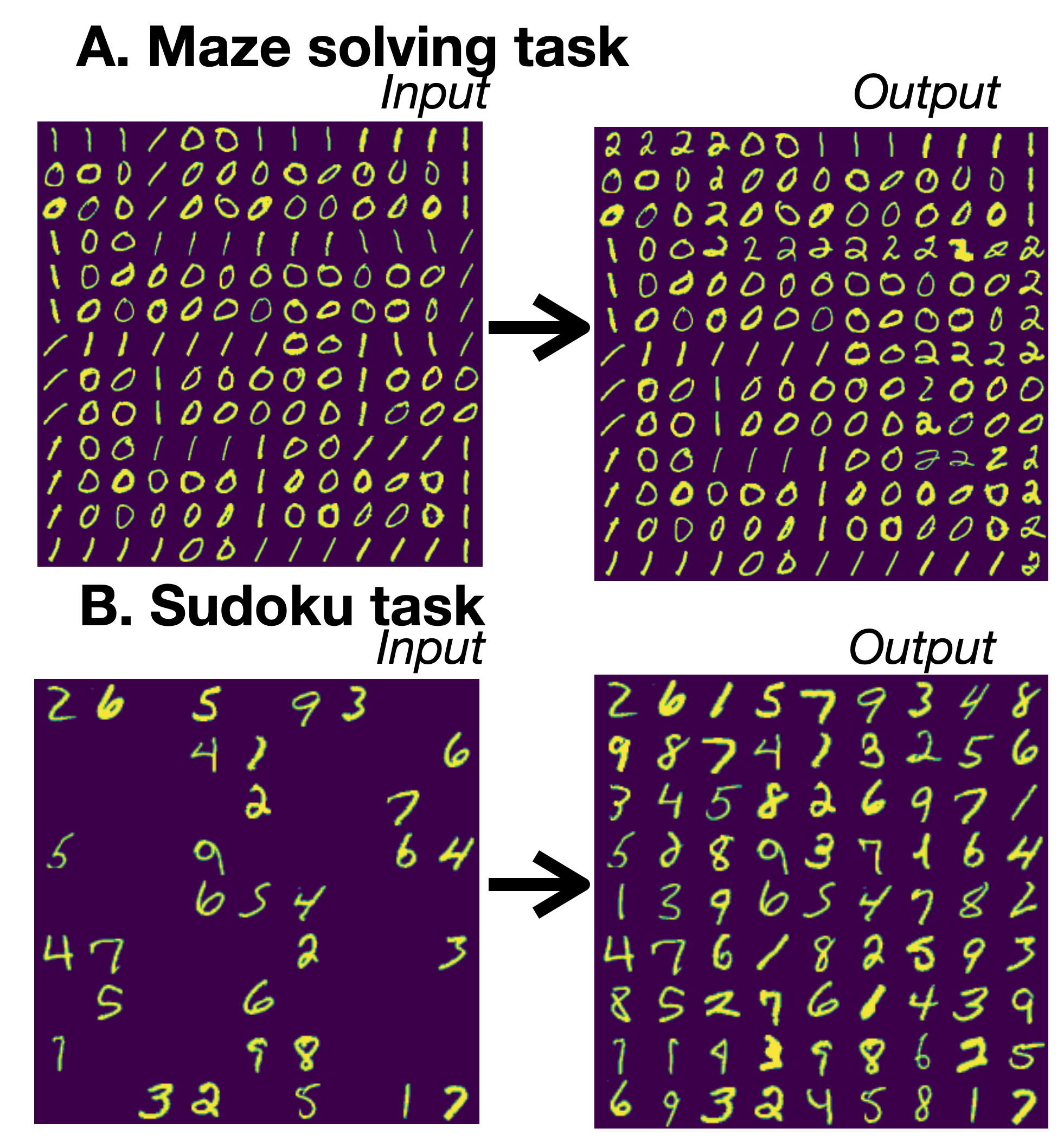}   
         \hspace*{0.05\linewidth}
          \includegraphics[width=0.52\linewidth]{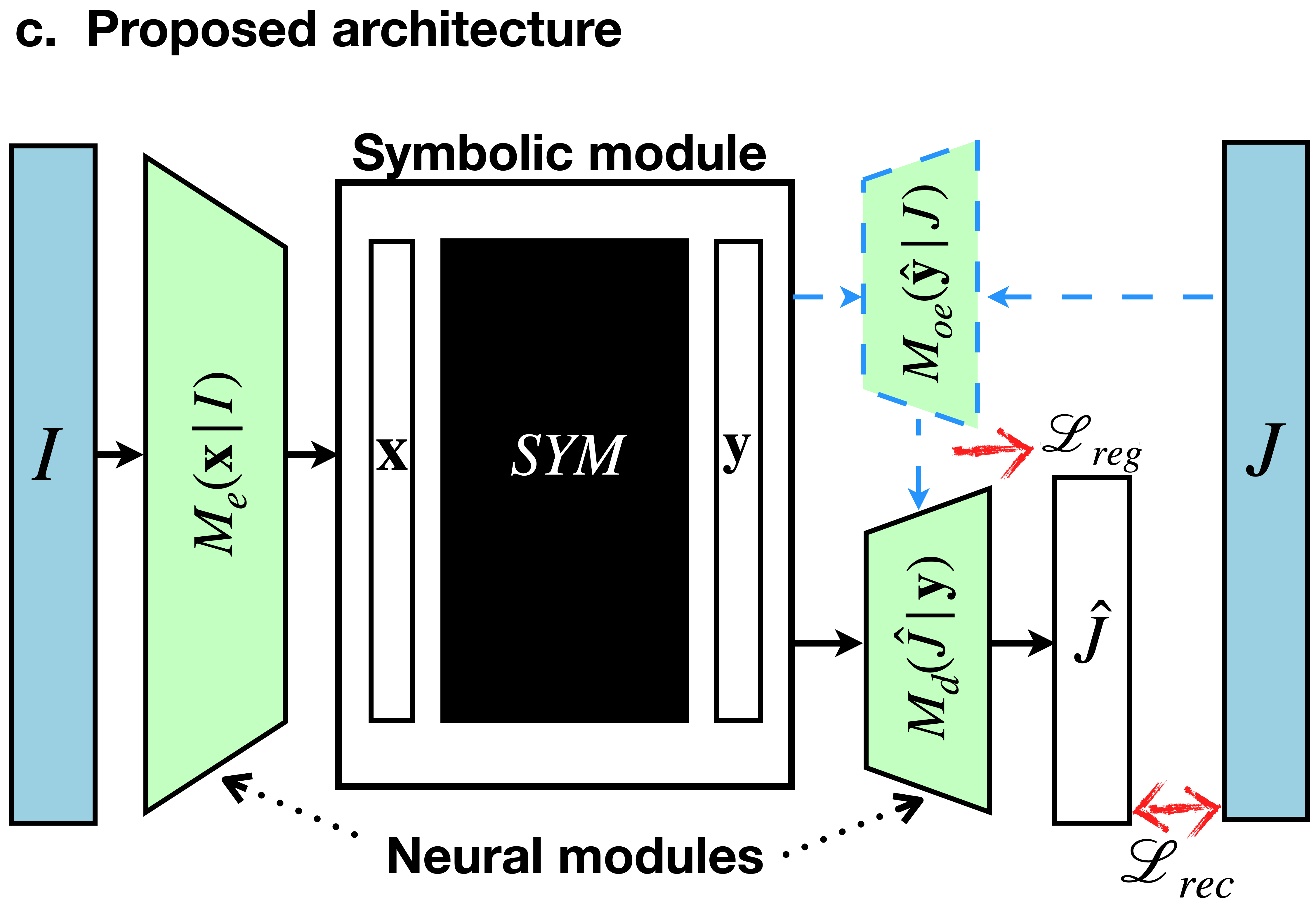}
    }
    \caption{(A,B) Two image-to-image reasoning tasks--maze solving, and sudoku solving, respectively-- where input and supervision are both in image form. (C) Schematic of \NSN, our neural-symbolic-neural system -- see Section~\ref{sec:formulation} for details. }
    \label{fig:architecture}
\end{figure}

We address this challenge via \NSN, our unique architecture and training routine (Figure~\ref{fig:architecture}C). The symbolic computation module (\emph{SYM}) is sandwiched between an encoder-decoder pair ($M_e, M_d$) that parse the input and produce output images respectively. $M_d$ provides a \textit{reconstruction loss} by reconciling \emph{SYM}'s output with the supervisory image. Simultaneously, the supervisory image is parsed and combined with \emph{SYM}'s output by a second encoder $M_{oe}$, used only during the training phase, to provide additional signal to $M_d$; this auxiliary \textit{regularization} significantly improves the quality of the learned reconstructions. We derive update equations to train the parameters of input encoder $M_e$ and show that they lead to a \textit{policy gradient}-like training procedure. This combination of enhanced policy gradient and regularization enables E2E training of the NSN framework.

We found that existing enhancements to policy gradient algorithms were insufficient for addressing the sparse reward distributions in our problem domains. To address this, we propose a \textit{subsampling heuristic} for approximating the policy gradient, and show that it significantly improves upon previous exploration strategies, and in some cases is necessary for learning in these tasks. We demonstrate our model's performance in solving the testbed problems in Figure~\ref{fig:architecture}A;B. We compare against purely neural architectures that work in an image-to-symbolic-solution setting (strictly weaker than our proposed setting) and demonstrate significantly better accuracy with much lesser training data. In particular, we show that encoding and task completion accuracy in \NSN\ closely tracks that of an ad-hoc system using pretrained classifiers and symbolic solvers, despite our image-to-image training paradigm and additional goal of generating output images. Taken together, we give a proof of concept on how to effectively build E2E hybrid NSN architectures for image-to-image reasoning tasks. We hope that our results form a basis for further research in this important area.

\section{Related work}
\textbf{Purely neural reasoning:} 
Recent works propose E2E neural architectures for classically symbolic computations and reasoning frameworks: e.g., satisfiability~\citep{Wang2019}, constraint satisfaction \citep{Nandwani2021}, inductive logic programming~\citep{Payani2019}, and value iteration~\citep{Tamar2016}. These models incorporate continuous (differentiable) relaxations of discrete algorithms and hence can be trained using backpropagation in E2E supervised fashion.
Although early results show promise, these architectures incur high training cost, perform approximate/black-box computations, and typically generalize poorly across problem sizes.

\textbf{Neural models with background knowledge:} Adding domain insights into data-trained neural models is a common pursuit in current ML systems. A popular approach for this is via \emph{constraints} in output space~\cite{Xu2018,Nandwani2019}. They are used in a separate inference step over the output of neural model, or converted to a differentiable loss added to the neural model's objective. These ideas have been applied in diverse domains such as sequence labeling in NLP \cite{Kolluru2020}, and recovering crystalline structure from multiple x-ray diffraction measurements of a material \cite{Chen2020}. Our proposal is similar in that we also provide additional domain knowledge to the model, but different, because we provide it in the form of an arbitrary (black-box) symbolic reasoner, which cannot be made differentiable. We also do not require pre-trained decoders unlike some prior work ( e.g., Chen et al.~\cite{Chen2020}).


\textbf{Neural parsing of visual input:} 
Recent work examines self-supervision in learning symbolic representations for visual input. Often, they are applied in video settings, e.g.,  where pixel similarities in consecutive frames in Atari games can help decode object identities \cite{Anand2019}, or the underlying object-layout map \cite{Kipf2020}. As with most unsupervised/self-supervised learning, there are no guarantees of learning a specific, complete symbolic parse of a scene. In contrast,  end-to-end accuracy in our setting heavily depends on {\em precisely} identifying and explicitly representing all symbols in the input, without which the symbolic computation may fail catastrophically.\footnote{In the testbeds we explore, we have assumed highly structured input formats; however, this is not a necessary part of the general architecture proposed in Figure~\ref{fig:architecture}.}


\textbf{Neural to symbolic architectures: } There is an emerging body of work in NS architectures, where neural components extract symbolic representations on which symbolic computations are performed. An example is visual question answering (QA) where textual questions are neurally converted to symbolic programs operating on representations extracted from an image \citep{Yi2018,Mao2019} -- they use a small, domain-specific language tailored towards benchmark tasks. 
Within textual QA, N2S models convert mathematical questions to mathematical expressions, which are directly executed to compute an answer \citep{Chen2020math}, and convert knowledge questions to a query that is run over a knowledge-base \citep{Liang2018}. Similar ideas are explored in other NLP tasks, e.g., dialog systems \citep{Raghu2021}. Probably closest to our work are N2S models that have been trained to perform perceptuo-reasoning, e.g., number addition, multiplication, where images represent the input, but outputs are symbolic (numeric) answers \citep{Tsamoura2021}. 

The main approaches to train such NS models include (1) the use of continuous relaxations of symbolic programs so that they can be differentiated end-to-end \citep{Manhaeve2018,Mao2019}, (2) a careful enumeration of \textit{possible inputs} (or programs) corresponding to an observed symbolic output, which can provide weak supervision \citep{Liang2018,Tsamoura2021}, and (3) pre-training neural components separately using intermediate supervision \citep{Yi2018}. In contrast our work differs in multiple ways. First, we build a novel \textit{neuro-symbolic-neuro} (NSN) architecture where input and supervision are both in the image domain and require training of two neural components, on either side of symbolic. Second, we aim to support black-box symbolic computations of arbitrary complexity (here, a graph algorithm and a constraint satisfaction program). Third, we handle highly structured intermediate and output spaces. Finally, we wish to train the whole model end-to-end, instead of using heuristic labeling or intermediate annotation to train components. To the best of our knowledge, this is the first demonstration of an E2E-trained NSN architecture.\footnote{see a recent compilation (Sarker et al.~\citep{Sarker2021} Figure 2) for a summary of existing neuro-symbolic architectures, which does not include NSN.}

\section{Problem formulation and approach}
\label{sec:formulation}
We are given a training dataset of the form $D=\{I^i, J^i\}_{i=1}^N$, where $I^i$ is an input image and  $J^i$ is a correct output image for $I^i$.  Our goal is to induce a function $f_\Theta$, such that $f_\Theta(I)=J$ only if $J$ is a correct output for image $I$. Our work focuses on tasks that require an intermediate reasoning step, for which a symbolic solver may be the most appropriate. So, we propose $f_\Theta$ as a novel NSN architecture -- a pipeline of neural, symbolic and neural components (see Figure \ref{fig:architecture}C).  The input $I$ is mapped by a neural encoder $\mathcal{M}_e({\bf \hat{x}} | \mathcal{I}; \Theta_e)$ to a distribution over symbols ${\bf \hat{x}}$, which in turn is proceesed by a symbolic program $\textit{SYM}(\mathbf{\hat{x}})$ to  output symbols ${\bf \hat{y}}$. Finally, a neural decoder  $\mathcal{M}_d(\mathbf{\hat{y}}; \Theta_d)$ generates output image $\hat{J}$ from supplied symbols. 

\begin{wrapfigure}[8]{r}{2.2in}
\vspace*{-2em}
    \includegraphics[width=2in]{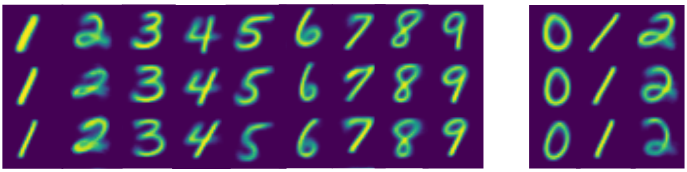}
    \caption{Sample digits generated by the maze solver (right) and sudoku model (left). More in appendix.}
    \label{fig:samples} 
\end{wrapfigure}

At training time, the only available feedback is the similarity between $\hat{J}$ and $J$. However, given only the symbol ${\bf \hat{y}}$, it is impossible to reconstruct $J$ exactly, due to potential variations in the representation $\hat{J}$ of ${\bf \hat{y}}$ (see Fig.~\ref{fig:samples}). We therefore need access to \emph{style information} ${\bf \hat{z}}$ implicit in $J$, and disentangled from its symbolic representation, allowing us to reconstruct $J$. We require another neural module $\mathcal{M}_{oe}({\bf \hat{y}}, J; \Theta_{oe})$ to extract style information ${\bf \hat{z}}$. The decoder $M_d$ must be modified to accept both ${\bf \hat{z}}$ and ${\bf \hat{y}}$, $\hat{J} = M_d({\bf\hat{y}}, {\bf\hat{z}}; \Theta_d)$. At testing time, since the true output $J$ is hidden, ${\bf\hat{z}}$ is sampled randomly. Since $M_{oe}$ is only relevant at training time, it is indicated with dotted lines in Fig.\ref{fig:architecture}.  The task requires optimizing $\Theta = \Theta_e\cup\Theta_d\cup\Theta_{oe}$ to induce the best $f_\Theta$, given only $D$ -- no intermediate supervision (data mapping $I$ to ${\bf x}$ or ${\bf y}$ to $J$, or style information ${\bf z}$) is made available to the model.


In our experiments, we consider highly structured input and output spaces (mazes and sudoku puzzles, Figure~\ref{fig:architecture}A;B), where each image is a grid of cells (sub-images). While our exposition of the \NSN\ architecture is general, for our problems, we implement $M_e$ and $M_d$ at the cell level. The exact architectural details of $M_e, M_{oe}, M_d$ are described in Sec.~\ref{sec:modeling}.


\subsection{Optimization objectives and update equations}

In the forward pass, the encoder $\mathcal{M}_e({\bf \hat{x}} | \mathcal{I}; \Theta_e)$  transforms input image $\mathcal{I}$ into a probability distribution over symbolic representations ${\bf \hat{x}}$, which in turn is processed by the symbolic algorithm \emph{SYM} to produce a symbolic output ${\bf \hat{y}}$. During training, output encoder $\mathcal{M}_{oe}({\bf \hat{y}}, J; \Theta_{oe})$ extracts style information ${\bf \hat{z}}$ from the ground truth output image $J$ \footnote{$M_{oe}$ is written as a deterministic function here for simplicity, however, it could in general also output a distribution over ${\bf \hat{z}}$}. The symbolic output is decoded by another model $\mathcal{M}_d({\bf \hat{y}}, {\bf \hat{z}}; \Theta_d)$ to produce a candidate output image $\hat{\mathcal{J}}$. This is compared to the supervisory image $\mathcal{J}$ to provide feedback to the models $\mathcal{M}_e(\cdot), \mathcal{M}_d(\cdot), \mathcal{M}_{oe}(\cdot)$. Therefore, we require the output encoder-decoder pair $\mathcal{M}_d, \mathcal{M}_{oe}$ to expose a \textit{reconstruction loss} function $l_{rec}({\bf \hat{y}}, \hat{J}, \mathcal{J} ; \Theta_d, \Theta_{oe})$ that compares generated image $\hat{J}$ with $J$ and returns a loss based on some measure of similarity between them. We can define the expected reconstruction loss used during training as
\begin{align}
\mathcal{L}_{rec}(I, J; \Theta_e, \Theta_d, \Theta_{oe}) = \mathop{\mathbb{E}}_{\mathcal{M}_e({\bf \hat{x}}\mid I; \Theta_e)} \left[{l}_{rec}({\bf \hat{y}}, \hat{J}, J; \Theta_d, \Theta_{oe})\right]
\end{align}
where $\hat{J} = \mathcal{M}_d({\bf \hat{y}}, {\bf \hat{z}}; \Theta_d)$, ${\bf \hat{z}} = \mathcal{M}_{oe}({\bf \hat{y}}, J; \Theta_{oe})$ and ${\bf \hat{y}} = \mathit{SYM}\left({\bf\hat{x}}\right)$. In addition to $l_{rec}$, the neural decoder may supply additional loss terms for regularization $l_{reg}({\bf \hat{y}}, {\bf \hat{z}}, \hat{J}, \mathcal{J} ; \Theta_d, \Theta_{oe})$. $l_{reg}$ is not directly related to quality of reconstruction and depends on the specific architecture being used for $M_d$. It is usually required to prevent overfitting or enforce certain properties. For instance, VAEs \cite{kingma2013auto} use a KL divergence term to enforce a continuous latent space (see Sec.~\ref{sec:modeling}) Accordingly, we define $L_{reg}$:
\begin{align}
\mathcal{L}_{reg}(I, J; \Theta_e, \Theta_d, \Theta_{oe}) = \mathop{\mathbb{E}}_{\mathcal{M}_e({\bf \hat{x}}\mid I; \Theta_e)} \left[l_{reg}({\bf \hat{y}}, {\bf \hat{z}}, \hat{J}, \mathcal{J} ; \Theta_d, \Theta_{oe})\right]
\end{align}
Putting together these two components, we have the overall loss for an example $(I, J)$
\begin{align}
\mathcal{L}(I, J) =  \mathcal{L}_{reg}(\mathcal{I},\mathcal{J}; \Theta_e, \Theta_d, \Theta_{oe}) + \mathcal{L}_{rec}(\mathcal{I},\mathcal{J}; \Theta_e, \Theta_d, \Theta_{oe}) \label{eq:loss}
\end{align}
We will now omit specifying parameters $\Theta_e, \Theta_d, \Theta_{oe}$ in subsequent equations for brevity. 

{\bf Training the neural modules:} We train the parameters $(\Theta_e, \Theta_d, \Theta_{oe})$ using supervision from the target image $\mathcal{J}$; in other words, the training procedure is performed end-to-end, with no access to supervision on the correct values of (internal) symbolic representations $({\bf x, y})$.  We can rewrite the expectation in the loss as
\begin{align*}
 \mathcal{L}(I, J) &= \mathop{\mathbb{E}}_{\mathcal{M}_e({\bf \hat{x}}\mid I)} \left[{l}_{rec}({\bf \hat{y}}, \hat{J}, J) + {l}_{reg}({\bf \hat{y}}, {\bf \hat{z}}, \hat{J}, J)\right] \\ 
 &= \sum_{{\bf \hat{x}}} \left({l}_{rec}({\bf \hat{y}}, \hat{J}, J) + {l}_{reg}({\bf \hat{y}}, {\bf \hat{z}}, \hat{J}, J)\right) \mathcal{M}_e({\bf \hat{x}}\mid I) \numberthis
\end{align*}
where the sum is taken over all possible ${\bf \hat{x}}$ and $\mathcal{M}_e({\bf \hat{x}}\mid I)$ is the probability of sampling ${\bf \hat{x}}$. Gradients with respect to the parameters $\Theta_e$ in the neural encoder can now be approximated using Monte-Carlo samples. Notice that ${\bf \hat{z}}$ and $\hat{J}$ only depend on $\Theta_d, \Theta_{oe}$ and therefore can be treated as scalar constants as far as the operation $\nabla_{\Theta_e}$ is concerned.  
\begin{align*}
\nabla_{\Theta_e}\mathcal{L}(I, J) &= \sum_{{\bf \hat{x}}} \left({l}_{rec}({\bf \hat{y}}, \hat{J}, J) + {l}_{reg}({\bf \hat{y}}, {\bf \hat{z}}, \hat{J}, J)\right)\cdot \nabla_{\Theta_e}\mathcal{M}_e({\bf \hat{x}}\mid I) \\
&= \sum_{{\bf \hat{x}}} \left({l}_{rec}({\bf \hat{y}}, \hat{J}, J) + {l}_{reg}({\bf \hat{y}}, {\bf \hat{z}}, \hat{J}, J)\right)\cdot \mathcal{M}_e({\bf \hat{x}}\mid I)\cdot  \nabla_{\Theta_e}\log\mathcal{M}_e({\bf \hat{x}}\mid I)  \\
&\approx \frac{1}{K}\sum_{i=1}^K \left(l_{rec}({\bf \hat{y}_i}, \hat{J}_i, J) + {l}_{reg}({\bf \hat{y}_i}, {\bf \hat{z}_i}, \hat{J}_i,  J)\right)\cdot\nabla_{\Theta_e}\log\mathcal{M}_e({\bf \hat{x}_i}\mid I) \numberthis 
\end{align*}
where the last step in Eq.~\ref{eq:pxme} is a Monte Carlo approximation of the sum by averaging $K$ iid samples $\left\{{\bf \hat{x}_i}\right\}_{i=1}^K$ drawn from the distribution given by $M_e$ and ${\bf \hat{z}_i}, \hat{J}_i, {\bf \hat{y}_i}$ are obtained in the usual way. Note that Eq.~\ref{eq:pxme} is structurally identical to the REINFORCE \cite{Williams1992} algorithm for calculating policy gradient. To draw parallels with a one-step RL problem (or equivalently, a contextual bandits problem), $l_{rec}({\bf \hat{y}}_i, \hat{J}_i, J) + {l}_{reg}({\bf \hat{y}_i}, {\bf \hat{z}_i}, \hat{J}_i,  J)$ corresponds to the reward, ${\bf \hat{x}}$ to the action and $\mathcal{M}_e({\bf \hat{x}}\mid I)$ to the policy with $I$ as the initial state (more details in appendix). However, updates to $\Theta_d, \Theta_{oe}$ cannot be made using policy gradients alone and a separate calculation is needed that yields a new term. Eq.~\ref{eq:pxmd} shows that updates to $\Theta_d$ follow conventional backpropagation rules, weighted by input distribution $M_e(I)$. 
\begin{align*}
\nabla_{\Theta_d}\mathcal{L}(I, J) &= \sum_{{\bf \hat{x}}} \nabla_{\Theta_d}\left({l}_{rec}({\bf \hat{y}}, \hat{J}, J) + {l}_{reg}({\bf \hat{y}}, {\bf \hat{z}}, \hat{J}, J)\right)\cdot \mathcal{M}_e({\bf \hat{x}}\mid I) \\
&\approx \frac{1}{K}\sum_{i=1}^K \nabla_{\Theta_d}\left(l_{rec}({\bf \hat{y}_i}, \hat{J}_i, J) + {l}_{reg}({\bf \hat{y}_i}, {\bf \hat{z}_i}, \hat{J}_i,  J)\right)\numberthis \label{eq:pxmd}
\end{align*}
where the last line is again a Monte Carlo approximation. Since $M_e$ only takes $\Theta_e$ as input, the probabilities $\mathcal{M}_e({\bf \hat{x}}\mid I)$ in Eq.~\ref{eq:pxmd} above are treated as constants with respect to the operation $\nabla_{\Theta_d}$. The update equations for $\Theta_{oe}$ are identical to Eq.~\ref{eq:pxmd}. Since Eq.~\ref{eq:pxmd} and its counterpart for $\Theta_{oe}$ may be viewed as expectated gradients of the reward function $l_{rec}({\bf \hat{y}_i}, \hat{J}_i, J) + {l}_{reg}({\bf \hat{y}_i}, {\bf \hat{z}_i}, \hat{J}_i,  J)$, we henceforth refer to these together as the \emph{reward gradient}. In practice, we find that when updating $\Theta_e$ it is beneficial to drop the $l_{reg}$ losses since they are primarily used to prevent decoder  overfitting and do not give useful information about the reconstruction quality. The gradient for $\Theta_e$ then becomes
\begin{align*}
\nabla_{\Theta_e}\mathcal{L}(I, J) 
\approx \frac{1}{K}\sum_{i=1}^K l_{rec}({\bf \hat{y}}_i, \hat{J}_i, J)\cdot\nabla_{\Theta_e}\log\mathcal{M}_e({\bf \hat{x}_i}\mid I) \numberthis \label{eq:pxme}
\end{align*}

\subsection{Subsampling Heuristic for Policy Gradient}
\label{sec:pg}

A challenge with policy gradient methods is the difficulty of learning in large, sparse-reward spaces, especially when reward distributions are non-smooth. Many previous approaches stress the need for effective exploration of the policy space, especially in the early stages of learning~\cite{Williams1991,Norouzi2016,Liang2018}; we sketch a couple below. Consider an RL problem with reward function $r$, policy $\pi_\theta$, initial state $s_0$ and prior over initial states $p$. Let $\pi_\theta(\textbf{a}\mid s_0)$ denote the probability of executing action sequence $\textbf{a}$ under $\pi_\theta$. The RL objective maximizes expected reward $\mathcal{O}_{RL} = \mathop{\mathbb{E}}_{p(s_0)} \left\{\sum_{\textbf{a}} \pi_\theta({\bf a}\mid s_0)r(\textbf{a}\mid s_0)\right\} $.
Williams \& Peng~\cite{Williams1991} propose adding an entropy-based regularization term $-\tau \pi_\theta(\textbf{a}|s_0) \text{log} \pi_\theta(\textbf{a}|s_0)$ to the objective $\mathcal{O}_{RL}$ in order to encourage exploration. Norouzi et al.~\cite{Norouzi2016} show that $\pi_\theta(\textbf{a}|s_0)$ maximizes this regularized expected reward, when it is directly proportional to the exponentially-scaled reward distribution, and propose to optimize a reward-scaled predictive probability distribution:
\begin{align}
    \mathcal{O}_{RAML} &=\mathop{\mathbb{E}}_{p(s_0)}\left\{ \tau\sum_{\textbf{a}} \pi^*_\tau(\textbf{a}| s_0)\text{log } \pi_\theta(\textbf{a}|s_0) \right\}
\end{align}
where $\pi^*_\tau(\textbf{a}|s_0) \propto \text{exp}\left\{ \frac{1}{\tau} r(\textbf{a}|s_0)\right\}$ represents a probability distribution over exponentiated rewards.  UREX~\cite{Nachum2017} uses a self-normalized importance sampling for Monte Carlo estimation of this objective, since it is often difficult or impossible to sample directly from the reward distribution, as is the case in our problem settings. They find that the UREX objective promotes the exploration of high-reward regions.  

In early experiments, we found that these exploration strategies did not work in our problem domains; the models were stuck in cold-start (Sec.~\ref{sec:ablations}). Instead, we propose to \textit{bias the gradient calculation} towards high-reward regions, by means of a \textit{subampling} algorithm for policy gradient. From the Monte Carlo samples used to calculate the expectation in $O_{RL}$, we keep only the top $\eta$ fraction of the samples, and discard the rest (see appendix for pseudocode). We refer to the combination of REINFORCE, the reward gradient and subsampling as the \NSN\ gradient. This builds on the ideas of reward-scaled policy gradient; low-reward samples are unlikely to have useful gradient information in sparse reward spaces, and therefore we discard them entirely. Although this leads to biased gradients, we hypothesize that in large action spaces the resulting bias-variance trade-off is favourable and mitigates the cold-start issue.\footnote{Although we do not consider it in this work, a suitable schedule such that $\eta \rightarrow 1$ can remove bias after overcoming cold start. We leave this as an open question for future work.} Recent work~\citep{Sinha2020} uses a similar approach in the context of training the generators in GANs, and finds that performance is significantly improved by discarding samples rated by the discriminator as less realistic.

\section{Experimental testbeds}
\label{sec:instantiations}

We evaluate our proposed framework in the context of two experimental testbeds: \vmaze\ and \vsud\  (Figure~\ref{fig:architecture}A,B). In the \vmaze\ task, the input is a maze where the walls and empty spaces are  samples of the digits ``1'' and ``0'' from the MNIST handwritten digits dataset.  The expected output is another image of the maze, with the cells on the shortest path from top-left to bottom-right are marked with samples of the digit ``2'' from the same dataset.  Mazes are designed such that there is a unique shortest path between these two points. In the \vsud\ task, the input is an unsolved (i.e., partially filled in) sudoku  with exactly one solution, and the digits 1-9 in each cell are, again, samples from the MNIST dataset. The expected output is the same puzzle correctly completed using a sampled instance of the correct digit in each cell. 

Other works have used similar settings: Tamar et al.~\cite{Tamar2016} evaluate their deep learning approach for value iteration on a shortest-path problem for mazes. Mulamba et al.~\cite{Mulamba2020} combine a neural classifier's predictions and uncertainty estimates both in a constraint satisfaction based solver for visual sudoku. Wang et al.~\cite{Wang2019} use sudoku to demonstrate their neural implementation of satisfiability solvers. In all these cases, the supervision (and network output) is symbolic in nature, and therefore they address only a portion of the problem we pose here: that of image-to-image learning. 


\subsection{Modeling details}
\label{sec:modeling}

Here we describe the implementations of neural modules $M_e, M_d$ and $M_{oe}$ that we use in our two tasks. Recall that our input and output images are a grid of cells, so apply encoder and decoder at cell-level, instead of image level. Let input image $I$ be a grid of known width $W$ and height $H$. Let $I_{w,h}$ represent the sub-image of a single digit at location $(w,h)$ in the grid. $\mathcal{M}_e(\hat{\textrm{x}}_{w,h} | \mathcal{I_{w,h}}; \Theta_e)$ computes the symbol $\hat{\textrm{x}}_{w,h}$, which are pieced together for all cells to output ${\bf \hat{x}}$. Similarly, ${\bf \hat{y}}$ comprises of $\hat{\textrm{y}}_{w,h}$ and ${\bf \hat{z}}$ of $\hat{\textbf{z}}_{w, h}$, each independently extracted from each sub-image $J_{w, h}$ as $M_{oe}(\hat{\textrm{y}}_{w, h}, J_{w, h}; \Theta_{oe})$. Each $\hat{\textrm{y}}_{w, h}$ is independently decoded to a sub-image $\mathcal{\hat{J}}_{w,h}$ as $\mathcal{M}_d( {\hat{\textrm{y}}_{w,h}}, {\hat{\textbf{z}}_{w,h}}; \Theta_d)$. $\hat{J}$ is constructed by putting together $\hat{J}_{w,h}$ in the grid. As an example, for Sudoku, the model applies the decoder on each cell of the 9x9 grid, and passes the classifier output and symbol layout to \emph{SYM}. Since \emph{SYM} has a well-specified input and output characteristic, the \textit{size of the symbol dictionary} is assumed to be known -- Sudoku puzzles contain 10 different symbols (9 digits and a blank space).  However, the mapping from visual input to symbol ID is not annotated; as a result, $M_e$ may learn permutations of the symbol dictionary in Sudoku, since the semantics of each digit are interchangeable.

\textbf{Input Encoder:} In the \vsud\ problem, we learn a LeNet \cite{lecun1989backpropagation} \textit{digit classifier} as part of our encoder model $M_e(\hat{\textrm{x}}_{w, h}|I_{w, h})$ -- given the contents of each cell, they classify the input as one of 10 digits (0-9). For the \vmaze\ task we use a fully connected network with a single hidden layer since only two digits need to be classified (0-1). 

\textbf{Output Encoder-Decoder:} The encoder and decoder models $M_d, M_{oe}$ in both tasks reconstruct each digit given a discrete label and ground truth image $J_{w, h}$. We choose an array of variational autoencoders (VAE) \cite{kingma2013auto} for the output encoder-decoder pair $\{V^i\}_{i=1}^B$ where $V^i$s are VAEs and there are $B$ of them. When generating output, the array is indexed by the symbolic input ${\rm \hat{y}_{w, h}}$ i.e. $V^{\rm \hat{y}_{w, h}}$ is selected to generate the image. 

Each $V^i$ consists of probability distributions $p_{\theta^i}$ and $q_{\phi^i}$ parametrized by $\theta^i$ and $\phi^i$ respectively. Image data $J$ is characterized by a space of latent variables ${\bf v}$ with joint probability distribution $p_{\theta^i}(J, {\bf v})$. The distribution $q_{\phi^i}({\bf v}\mid J)$ is optimized to approximate the posterior $p_{\theta^i}({\bf v}\mid J)$. The evidence lower bound (ELBO) for $V^i$ is 
\begin{align}
    \mathcal{L}_{\theta^i, \phi^i}(J) = \log p_{\theta^i}(J) - D_{KL}(q_{\phi^i}({\bf v}\mid J)\| p_{\theta^i}({\bf v} \mid J)) 
\end{align}
During training the ELBO is maximized, and therefore the loss function is $-\mathcal{L}_{\theta^i, \phi^i}(J)$. The first term in the loss, $-\log p_{\theta^i}(I)$ can be interpreted as a measure of quality of reconstruction, whereas the second $D_{KL}(q_{\phi^i}({\bf v}\mid J)\| p_{\theta^i}({\bf v} \mid J)) $ is a regularization term that forces $q_{\phi^i}({\bf v}\mid J) \approx p_{\theta^i}({\bf v}\mid J)$. Latent variables $\textbf{v}$ encode style information ${\bf \hat{z}_{w, h}}$. We therefore have $M_{oe} := \{q_{\phi^i}\}_{i=1}^B$ and $M_d := \{p_{\theta^i}\}_{i=1}^B$ along with the following expressions for $l_{reg}$ and $l_{rec}$. 
\begin{align}
    l_{rec}({\rm \hat{y}_{w, h}}, \hat{J}_{w, h}, J_{w, h}) &= -\log p_{\theta^{\rm \hat{y}_{w, h}}}(J_{w, h}) \\
    l_{reg}({\rm \hat{y}_{w, h}}, {\bf \hat{z}_{w, h}}, \hat{J}_{w, h}, J_{w, h}) &= D_{KL}(q_{\phi^{\rm \hat{y}_{w, h}}}({\bf \hat{z}_{w, h}}\mid J_{w, h})\| p_{\theta^{\rm \hat{y}_{w, h}}}({\bf \hat{z}_{w, h}} \mid J_{w, h})) 
\end{align}
where the parameters $\Theta_d = \bigcup_{i=1}^B\theta^i$ and $\Theta_{oe} = \bigcup_{i=1}^B\phi^i$. In the \vsud\ task the size of the VAE array is $B=9$, whereas in the \vmaze\ task it is $B=3$.

\textbf{Symbolic Algorithm:} For the \vmaze\ problem, ${\bf \hat{x}}$ represents a maze as a 2D array of labels (\{\textit{wall, empty}\}) for each cell, generated by running the  $M_e$ on each cell and independently sampling each label. \emph{SYM} is Dijkstra's shortest path algorithm, which finds a shortest path from source to the closest reachable point to the sink, and marks the shortest-path cells to create  the symbolic representation ${\bf x}\rightarrow{\bf y}$. The shortest path is not guaranteed to exist in each sample, even if it exists in the ground truth; in such cases, \emph{SYM} finds a path to the closest reachable point to the sink.  By design, we do not have access to the correspondence between input digits and labels; this correspondence, in addition to the classifier  itself, is learned via supervision from the image $J$.

For the \vsud\ problem, similar to the above procedure, the symbolic input ${\bf\hat{x}}$, is a sampled symbol from $M_e$'s output distribution on each filled-in cell, with the \{\textit{empty}\} symbol everywhere else. When sampling symbols to generate the input, we place a well-formedness constraint: we follow a fixed order, and resample the next cell until we produce a symbol that does not match any already sampled symbol in the same row or column (this is part of the Sudoku constraint set). We note that despite this partial well-formedness constraint, only a very small fraction of samples ${\bf\hat{x}}$ are solvable in the early stages of training. \emph{SYM} uses a constraint solver (see appendix) to solve the Sudoku and generate the missing digits for the output representation ${\bf\hat{y}}$. It is possible that the sampled Sudoku does not have a solution, in which case the solver fails and both $l_{reg}, l_{rec}$ are identically zero.

\section{Results}

\subsection{Comparison and metrics}
 We use two metrics to measure the performance of \NSN\ at end-to-end training: {\em task completion rate} (TCR), i.e., fraction of mazes / sudoku puzzles where the exact symbolic solution was found by the E2E system, and {\em classification accuracy} (CA), where we measure the fraction of input symbols correctly identified. \footnote{Since our system receives no intermediate feedback, \NSN\ has no knowledge of the ordering of symbols, for example, that slot 1 should correspond to the symbol \texttt{zero}. It therefore identifies symbols accurately upto a permutation; for measuring classification accuracy, we use the appropriate permutation for each trained model. For \vmaze, however, each symbol has a distinct role, necessary for the path solver, and so the network identifies symbols in the correct ordering. }
Since to the best of our knowledge, no current system accomplishes image-to-image training, we instead use purely neural image-to-symbolic models as our external baselines -- neural semidefinite programming for sudoku (\SATNET~\cite{Wang2019}), and value iteration networks for maze (\VIN~\cite{Tamar2016}). These baselines are an overestimate, since training them as-is in an NSN setting will yield worse performance, and potentially even be infeasible. We also compare against a $M_e$ that is pretrained on dataset $\{I^i_{w, h}, {\rm x}_{w, h}^i\}_{i=1}^N$ where ${\rm x}_{w, h}^i$ is the true label of digit $I^i_{w, h}$. This represents an upper bound on \NSN\ accuracy  which has no access to true labels. 
Finally, we perform ablations to highlight the importance of the subsampling heuristic and also compare 
against existing RL approaches designed for sparse reward spaces.


\subsection{Visual maze solving}

\begin{wraptable}[14]{r}{3.2in}
\vspace*{-3em}
    \begin{tabular}{cccccc}
    \toprule 
         \multirow{2}*{\textbf{Maze dim.}} & \multicolumn{2}{c}{\textbf{\NSN}} & \multicolumn{1}{c}{\textbf{\VIN}} & 
         \multicolumn{2}{c}{\textbf{Pretrained $\mathcal{M}_e$}} \\
         \cmidrule(lr){2-3} 
         \cmidrule(lr){4-4}
         \cmidrule(lr){5-6}
         & CA & TCR & TCR & CA & TCR \\ 
         \midrule
         $7\times 7$ & 99.95 & 99.4 & 96.9 & 99.96 & 99.7 \\
         $13\times 13$ & 99.91 & 99.1 & 29.2 & 99.95 & 99.2 \\ 
         $19\times 19$ & 99.96 & 97.7 & 4.4 & 99.95 & 98.4 \\ 
         $28\times 28$ & 99.81 & 95.5 & - & 99.95 & 96.8 \\ 
         $43\times 43$ & 99.89 & 93.9 & - & 99.95 & 93.5 \\
        \bottomrule
    \end{tabular}%
     \caption{Results on \vmaze\ for different maze sizes. \NSN\ (ours) classifies input symbols accurately (digit accuracy DA) and has high path accuracy (PA), even as maze sizes increase; \VIN\ scales poorly in contrast. Our results are close to that of a system using pretrained classifiers.} \label{tab:vmaze}
\end{wraptable}

\textbf{Data and training:}
We generated mazes by constructing random spanning trees on $n$x$n$ grids, with grid points as walls and edges between as empty spaces. This guarantees a unique shortest path between any two points; for our experiments we only used the top left cell, and the bottom right cell, as entry and exit points. We tested a range of maze sizes but not solution complexity or multiplicity; we anticipate no significant qualitative change in the results under those manipulations. Hyper-parameters for each method were determined by a grid search over batch size $\in \left\{25, 50, 100, 250\right\}$, learning rate $\in \left\{10^{-3}, 10^{-4}\right\}$. For subsampling, we tried fraction $\eta \in \left\{1, \frac{1}{2}, \frac{1}{16}, \frac{1}{32}, \frac{1}{128}\right\}$ while the number of samples $K$ was fixed at 128. For \VIN, the number of value iterations was chosen to be slightly larger than the maze dimension. All experiments using \NSN\ were run for 80 epochs whereas for \VIN\ we used early stopping based on a dev set since it is prone to overfitting when the dataset is small. We do not report numbers on large mazes for \VIN, since train time becomes extremely large and TCR has already dropped to $<5\%$ for $19\times 19$. The train, dev and test set each contained 1000 maze problems for each grid size.  




\textbf{Results:} Table~\ref{tab:vmaze} shows the performance of \NSN\ on the \vmaze\ problem as the size of the mazes are varied. Classifier accuracy remains high even as mazes grow large.  We also see that while task completion rate (path accuracy) remains high, small changes in classifier accuracy leads to large degradations in task completion rates; this is because we consider a 0-1 metric for task completion, where all path cells need to be correctly identified, partially correct paths are not rewarded.

\begin{wraptable}[12]{r}{2.7in}
\vspace*{-2em}
    \begin{tabular}{ccccc}
    \toprule 
         \multirow{2}*{\textbf{Difficulty}} & \multicolumn{2}{c}{\textbf{\NSN}} &  
         \multicolumn{2}{c}{\textbf{Pretrained $\mathcal{M}_e$}} \\
         \cmidrule(lr){2-3} 
         \cmidrule(lr){4-5}
         & CA & TCR & CA & TCR \\ 
         \midrule
         Easy & 98.56 & 65.4 & 98.70 & 68.9 \\
         Medium & 97.14 & 62.8 & 98.27 & 69.4 \\ 
         Hard & 96.52 & 61.0 & 97.49 & 70.0 \\ 
        \bottomrule
    \end{tabular}
    \label{tab:my_label}
     \caption{Results on \vsud\ tasks for different puzzle hardness levels. \NSN\ shows high input symbol classification accuracy (CA) and task completion rate (TCR) across levels.}\label{tab:sudoku}
\end{wraptable}

\subsection{Visual Sudoku solving}

\textbf{Data and training:}
Sudoku problems are generated by starting with a random $9\times 9$ grid $G$ that satisfies sudoku constraints. A problem $P$ is constructed by performing a local search over the space of sudoku problems with $G$ as a unique solution. The difficulty of $P$ is estimated by computing a linear combination of the square of branching factors $B_{ij}$ at each cell and the number of empty cells $E$ ( details in appendix). In general, problems of higher difficulty have fewer filled cells. 
Note that the symbolic constraint solver \emph{SYM} always gives the correct solution regardless of problem difficulty. We report numbers on three difficulty levels - easy, medium and hard (details in appendix on how these are defined). Train, test and val datasets contain 1000 sudoku puzzles. The best hyperparameters were determined by a grid search as in the \vmaze\ setting.

\textbf{Results:}
Table~\ref{tab:sudoku} shows results on \vsud\, demonstrating high classification accuracy and task completion rates across problem hardness. In particular, despite the image-to-image training, and additional challenge of output image generation, our performance is close to that of a system using pre-trained classifiers for parsing the input. 
Again, we see a strongly nonlinear relationship between classifier accuracy and task completion, with even very high classifier accuracies insufficient for proportionately high 0/1 task-completion rates. This is because a classifier must identify \textit{all} input digits in the Sudoku correctly to get credit. Finally, Wang el al. \cite{Wang2019} reports an accuracy of 63.2\% on their purely neural image-to-symbolic Sudoku solver \SATNET{}, using 9x our training data.

\begin{wraptable}[13]{r}{2.5in}
\vspace*{-2em}
    \begin{tabular}{ccccc}
    \toprule 
         \multirow{2}*{Fraction} &
         \multicolumn{2}{c}{\textbf{13x13 Maze}} & \multicolumn{2}{c}{\textbf{Sudoku}} \\
          \cmidrule(lr){2-3} 
         \cmidrule(lr){4-5}
         & CA & TCR & CA & TCR \\
        \midrule
        1 & 56.80 & 0.0 & 11.77 & 0.0 \\
        1/2 & 99.94 & 98.2 & 20.92 & 0.0 \\  
        1/16 & 99.95 & 99.2 & 38.65 & 0.0 \\
        1/32 & 99.81 & 98.0 & 98.35 & 60.9  \\ 
        1/128 & 99.91 & 99.1 & 98.56 & 65.4  \\
        \bottomrule
    \end{tabular}
    \caption{Accuracies for \vsud\ and \vmaze\ ($13\times13$) with varying subsampling fraction. Accuracy significantly increases when this fraction is lowered.}\label{tab:subsamp}
\end{wraptable}

\subsection{Ablations} 
\label{sec:ablations}
We try various values of subsampling fraction $\eta$ and observe its effect on accuracy (Table~\ref{tab:subsamp}) for \vsud\ and \vmaze\ on $13\times 13$ mazes. For $\eta=1$ the classifier accuracy is no better than chance and accuracy significantly increases as $\eta$ is lowered. Note that $\eta=\frac{1}{K}=\frac{1}{128}$ is the lowest value of $\eta$ where we pick just one sample. This shows that subsampling is indeed effective at solving the cold-start problem. 

We try various policy gradient alternatives to the \NSN\ gradient. One of these simply omits the reward gradient term (Eq.~\ref{eq:pxmd}). This method is not able to get non-zero TCR on either task,  illustrating that training requires supervision from the decoder in the form of reconstruction loss.
We also try REINFORCE with maximum entropy regularization \cite{Williams1991} and UREX \cite{Nachum2017}, adding a reward gradient term to each of these and find that they also do not give non-zero TCR on both tasks (detailed tables in appendix). This is because they cannot overcome the cold start problem as discussed in Sec.~\ref{sec:pg}. 

All experiments were run on a cloud VM based on the Intel Xeon Scalable Processor with 60 vCPUs and 240 GB memory and an internal cluster containing nodes with 40 Intel Xeon Gold CPUs and 100 GB memory. No GPUs were used since workloads were primarily CPU intensive because of the symbolic module \emph{SYM}. Each experiment was restricted to using atmost 12 threads.  

\section{Discussion}
 
We proposed a novel learning paradigm of image-to-image training involving symbolic reasoning over the contents of the images. We designed \NSN, an architecture for solving such problems; derived the learning rules for our system as an extension of policy-gradient-style algorithms; proposed and implemented optimizations to policy gradient to enable it to succeed on our learning problems. Finally, we showed via experimentation on two testbed problems the efficacy of our end-to-end training paradigm, and the advantage obtained, by design, over purely neural systems.
 \\
 Many lines of work in deep learning involve symbolic manipulation and processing of perceptual inputs, whether via implicit representations (``purely neural'' systems), or by explicit extraction and processing of symbolic inputs (neuro-symbolic systems).  We argue that our proposed learning paradigm is a natural next step in the evolution of hybrid neuro-symbolic architectures, where both input and supervision are obtained from naturalistic sources such as images. The broad promise of this line of research is the ability to train agents and AI systems with less dependence on annotation, intermediate supervision, or large amounts of training data. In addition, systems that maintain and manipulate explicit symbolic representations have the powerful advantages of interpretability and verifiability -- both properties that are increasingly necessary as we design complex AI systems.
 \\
 In future work, we aim to examine broader problems in this NSN space, extending our work to naturalistic problems, and towards developing general principles of blending neural and symbolic computations in a seamless manner.
 
\label{sec:notes}
\textbf{Limitations of our work:} This is one of many possible new directions for neuro-symbolic architectures that blend neural and symbolic components. Also, we chose simple, controlled testbeds for prototyping purposes. We expect the design, implementation, and real-world applicability of such systems will need to be refined through future work.
\\
\textbf{Societal impact:} We present an abstract/conceptual proposal for learning systems.  The intent and broad direction of our effort is to make learning systems more expressive and easier to train; an expected positive outcome. As noted above under limitations, the real-world value of our work is still to be determined and difficult to anticipate. We see no immediate ethical concerns of our work.

\bibliographystyle{unsrt}
\bibliography{main.bib}

\begin{appendices}
\section{Policy Gradient}
We describe the policy gradient approach in the original reinforcement learning formulation~\citep{Williams1992}, and various recent approaches; subsequently, we make connections to our learning problem. In policy gradient methods, the goal is to learn an \textit{action policy} for an agent, $\pi_\theta(a_t|\textbf{s})$ -- a probability distribution over actions $a_t$ associated with a sequence of states $\textbf{s} = (s_0, s_1, \ldots, s_{t})$ -- in order to maximize reward.  In classical RL formulations, states evolve in Markov fashion, conditioned on the agent's action at each state, and starting from some initial condition $s_0$, so we can calculate $\pi_\theta({\bf a}\mid s_0)$, the probability of the action sequence ${\bf a} = (a_0, \ldots, a_t)$ given initial state $s_0$ 
\begin{align}
\pi_\theta({\bf a}\mid s_0) = \mathbb{P}\left[\textbf{a} \mid s_0, \pi_\theta\right] = \prod_{t=0}^T \pi_\theta(a_t\mid s_t)
\end{align}
where the states $s_{t+1} = f(s_t, a_t)$ evolve according to some transition function $f(\cdot)$. 
The \reinf\ objective equals the expected reward when following a policy $\pi_\theta$
\begin{align*}\label{eq:RL}
    \mathcal{O}_{RL} &= \mathop{\mathbb{E}}_{s_0 \sim p} \left\{\sum_{\textbf{a}} \pi_\theta({\bf a}\mid s_0)r(\textbf{a}\mid s_0)\right\} \numberthis\\
    \nabla_\theta\mathcal{O}_{RL} &= \mathop{\mathbb{E}}_{s_0\sim p}\left\{ \sum_{\textbf{a}} r(\textbf{a}\mid s_0) \nabla_\theta \pi_\theta({\bf a}\mid s_0) \right\} \\
    &= \mathop{\mathbb{E}}_{s_0\sim p}\left\{ \sum_{\textbf{a}} \pi_\theta({\bf a}\mid s_0) r(\textbf{a}|s_0) \nabla_\theta \text{log}\pi_\theta({\bf a}\mid s_0) \right\} \numberthis
\end{align*}
where $p$ is a prior over initial states and $r$ is a reward function. The last sum is derived using the identity $\nabla_\theta f(\theta) = f(\theta) \nabla_\theta\log f(\theta)$. \cite{Williams1992} proposed Monte Carlo sampling to estimate this gradient, by averaging weighted rewards over samples drawn from the \textit{current policy}, i.e., over initial states $s_0^{(n)}\sim p(s_0)$, and action sequences $\textbf{a}^{(k)} \sim \pi_\theta(\textbf{a}|s_0^{(n)})$: 
\begin{align}
 \nabla_\theta \mathcal{O}_{RL} \approx \frac{1}{NK}\sum_{n=1}^{N} \sum_{k=1}^{K}  r(\textbf{a}^{k}|s_0^{(n)})\nabla_\theta\text{log} \pi_\theta(\textbf{a}^{k}|s_0^{(n)})  \label{eq:pg}
\end{align}
This equation is structurally similar to Eqs.~\ref{eq:pxme}, where the model $M_e$ and the symbolic representations ${\bf \hat{x}}$ correspond to the policy and actions, respectively, and the loss can be considered as a reward with a sign change.  The outer sum in the above equation is simply the aggregation of the pointwise gradient over training instances, to compute batch gradient. We therefore use policy gradient for learning the parameters of the encoder network $M_e$. The reward gradient (Eq.~\ref{eq:pxmd}) is an additional term that must be included to train $M_d$ and $M_{oe}$ as the policy gradient makes no updates to their parameters.  

\section{Pseudocode}
We describe the algorithm (algorithm~\ref{alg:RLsub}) for training a general \NSN{} architecture. This shows how to compute the reward gradient and update $\Theta_d, \Theta_{oe}$, and also how subsampling is used. 

\begin{center}
\begin{algorithm}[H]
  \caption{Training \NSN{}}
  \label{alg:RLsub}
\begin{algorithmic}
  \STATE {\bfseries Input:} data $\left\{I_i, J_i\right\}_{i=1}^N$, neural modules $\mathcal{M}_e$, $\mathcal{M}_d$, $M_{oe}$, number of samples $K$, learning rate $\alpha$, symbolic function $SYM$, subsampling fraction $\eta$, baseline $\beta$
  \REPEAT
  \STATE Divide data $I_i, J_i$ into batches $B$
  \FOR{batch $b$ in $B$}
  \STATE Initialize $\Delta_e \leftarrow 0, \Delta_d \leftarrow 0, \Delta_{oe}\leftarrow 0$
  \FOR{$I, J$ in $b$}
  \STATE Initialize empty array $A$, $\delta_e \leftarrow 0, \delta_d \leftarrow 0$, $\delta_{oe}\leftarrow 0$
  \STATE Let $\pi_{\theta} \leftarrow \mathcal{M}_e(I)$ be a probability distribution
   
  \FOR{$i=1$ to $K$}
  \STATE Draw random sample ${\bf \hat{x}}$ from $\pi_\theta$ with probability $\pi_\theta\left({\bf \hat{x}}\right)$
  \STATE Compute ${\bf \hat{y}} = SYM({\bf \hat{x}})$
  \STATE Extract style information ${\bf \hat{z}} = M_{oe}\left({\bf\hat{y}}, J\right)$
  \STATE Generate output image $\hat{J} = M_d\left({\bf\hat{y}}, {\bf\hat{z}}\right)$
  \STATE Append $\left\{{l}_{rec}({\bf \hat{y}}, \hat{J}, J), l_{reg}({\bf \hat{y}}, {\bf \hat{z}}, \hat{J}, \mathcal{J} ; \Theta_d, \Theta_{oe}), \pi_\theta\left({\bf \hat{x}}\right)\right\}$ to $A$
  \ENDFOR
  \STATE Sort $A_R$ in lexicographically decreasing order
  \STATE $K'\leftarrow \lfloor K\eta\rfloor$
  \FOR{$r, g, p$ in $A_R[1..M']$}
  \STATE $\delta_e \leftarrow \delta -  \left[(-r)-\beta\right]\cdot \nabla_{\Theta_e}\log p$
  \STATE $\delta_d \leftarrow \delta_d + \nabla_{\Theta_d}\left(r + g\right)$
  \STATE $\delta_{oe} \leftarrow \delta_{oe} + \nabla_{\Theta_{oe}}\left(r + g\right)$
  \ENDFOR
  \STATE $\Delta_e \leftarrow \Delta_e + \frac{\delta_e}{K'}$ 
  \STATE $\Delta_d \leftarrow \Delta_d + \frac{\delta_d}{K'}$
  \STATE $\Delta_{oe} \leftarrow \Delta_{oe} + \frac{\delta_{oe}}{K'}$
  \ENDFOR
  \STATE $\Theta_e \leftarrow \Theta_e - \alpha\frac{\Delta_e}{|B|}$ 
  \STATE $\Theta_d \leftarrow \Theta_d - \alpha\frac{\Delta_d}{|B|}$
  \STATE $\Theta_{oe} \leftarrow \Theta_{oe} - \alpha\frac{\Delta_{oe}}{|B|}$
  \ENDFOR
  \UNTIL{convergence}
\end{algorithmic}
\end{algorithm}
\end{center}

\section{Sudoku Constraint Solver}

We use \href{https://developers.google.com/optimization}{Google OR-Tools}, an open source library for optimization to solve Sudokus. In particular, we use the CP-SAT Solver which can be used to create and solve constraint satisfaction problems. OR-Tools is available under the Apache license. 

\section{Sudoku Generator}

We use code available on \href{https://dlbeer.co.nz/articles/sudoku.html}{Daniel Beer's blog} to generate Sudokus of varied difficulty levels. 

\subsection{Estimating difficulty scores}

Difficulty scores are estimated by looking at the branching factors $B_{ij}$ at empty cells $(i, j)$ that the Sudoku solver encounters at each step of the path from the root of the search tree to the solution. If $E$ is the set of empty cells then the difficulty score is estimated as

\begin{equation}
    100\cdot \sum_{(i, j) \in E} \left(B_{ij}-1\right)^2 + |E|
\end{equation}

Note that the difficulty score is simply $|E|$ for a problem that has $B_{ij} = 1$, i.e. requires no backtracking. In general, problems with scores less than $100$ are easy and require no backtracking and scores in the $300+$ range are fairly challenging. In Table~\ref{tab:sudoku}, easy puzzles correspond to scores of roughly $100$, medium to scores of $400$ and hard ones to scores of $600$.   

\subsection{Generating Sudokus}

As mentioned previously, Sudoku problems are generated by starting with a random $9\times 9$ grid $G$ that satisfies Sudoku constraints. $G$ is generated by a random backtracking search over $\{1, 2, \ldots, 9\}^{9\times 9}$, the space of $9\times 9$ grids of single digit positive integers. A problem $P$ is obtained from $G$ by clearing certain cells, ensuring that $G$ is the only solution of $P$. This is done by a local search - cells are randomly cleared from the grid and it is checked if $G$ is the only possible solution to the new problem. If this is the case, then the difficulty score is estimated and the problem is stored in the dataset.   

\section{Digit Samples}
Table~\ref{tab:maze_samples} includes samples for the symbols 0, 1 and 2 for mazes of different dimensions. Table~\ref{tab:sudoku_samples} includes samples for digits 1-9 for Sudoku problems of different hardness levels. Note that in table~\ref{tab:sudoku_samples} the digits do not correspond to the symbols since \NSN{} has no way of knowing the correspondence between symbols and images due to the lack of intermediate supervision. Therefore, in general, some permutation of symbols to MNIST images is learnt by the classifier and VAEs. 
\subsection{\vmaze} 
\newcommand{\sampleImageWidth}{0.15\textwidth}

\begin{table}[h!]
    \centering
    \begin{tabular}{cccc}
    \toprule
        \textbf{Dimension} & \textbf{0} & \textbf{1} & \textbf{2} \\
        \midrule
        $7\times 7$ & \includegraphics[width=\sampleImageWidth]{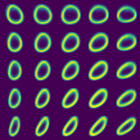} & \includegraphics[width=\sampleImageWidth]{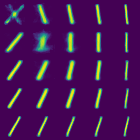} & \includegraphics[width=\sampleImageWidth]{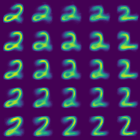} \\ 
        $13\times 13$ & \includegraphics[width=\sampleImageWidth]{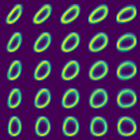} & \includegraphics[width=\sampleImageWidth]{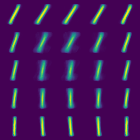} & \includegraphics[width=\sampleImageWidth]{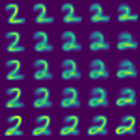} \\ 
        $19\times 18$ & \includegraphics[width=\sampleImageWidth]{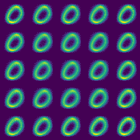} & \includegraphics[width=\sampleImageWidth]{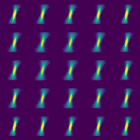} & \includegraphics[width=\sampleImageWidth]{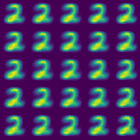} \\ 
        $28\times 28$ & \includegraphics[width=\sampleImageWidth]{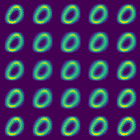} & \includegraphics[width=\sampleImageWidth]{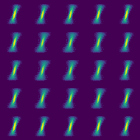} & \includegraphics[width=\sampleImageWidth]{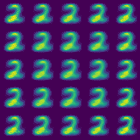} \\
        $43\times 43$ & \includegraphics[width=\sampleImageWidth]{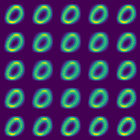} & \includegraphics[width=\sampleImageWidth]{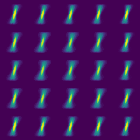} & \includegraphics[width=\sampleImageWidth]{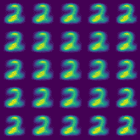} \\ 
        \bottomrule
    \end{tabular}
    \caption{Sample images for \vmaze\ }
    \label{tab:maze_samples}
\end{table}
\pagebreak
\subsection{\vsud}

\newcommand{\centered}[1]{\begin{tabular}{l} #1 \end{tabular}}

\begin{table}[h!]
    \centering
    \begin{tabular}{cccccccccc}
    \toprule 
            \textbf{Symbol} & \textbf{Easy} & \textbf{Medium} & \textbf{Hard} \\
    \midrule 
       1 & \includegraphics[width=\sampleImageWidth]{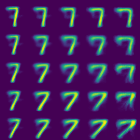} & \includegraphics[width=\sampleImageWidth]{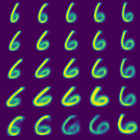} & \includegraphics[width=\sampleImageWidth]{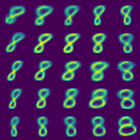} \\ 
        2 & \includegraphics[width=\sampleImageWidth]{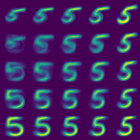} & \includegraphics[width=\sampleImageWidth]{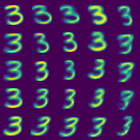} & \includegraphics[width=\sampleImageWidth]{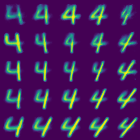} \\ 
        3 & \includegraphics[width=\sampleImageWidth]{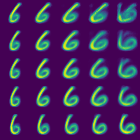} & \includegraphics[width=\sampleImageWidth]{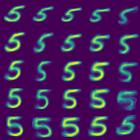} & \includegraphics[width=\sampleImageWidth]{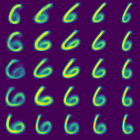} \\ 
        4 & \includegraphics[width=\sampleImageWidth]{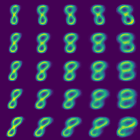} & \includegraphics[width=\sampleImageWidth]{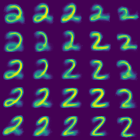} & \includegraphics[width=\sampleImageWidth]{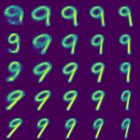} \\ 
        5 & \includegraphics[width=\sampleImageWidth]{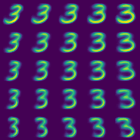} & \includegraphics[width=\sampleImageWidth]{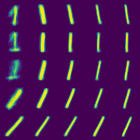} & \includegraphics[width=\sampleImageWidth]{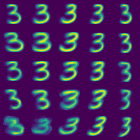} \\ 
        6 & \includegraphics[width=\sampleImageWidth]{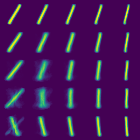} & \includegraphics[width=\sampleImageWidth]{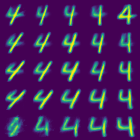} & \includegraphics[width=\sampleImageWidth]{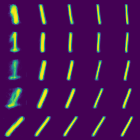} \\ 
        7 & \includegraphics[width=\sampleImageWidth]{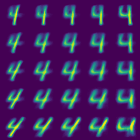} & \includegraphics[width=\sampleImageWidth]{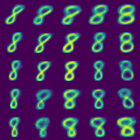} & \includegraphics[width=\sampleImageWidth]{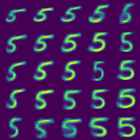} \\ 
        8 & \includegraphics[width=\sampleImageWidth]{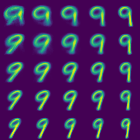} & \includegraphics[width=\sampleImageWidth]{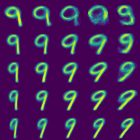} & \includegraphics[width=\sampleImageWidth]{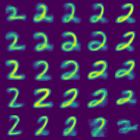} \\
        9 & \includegraphics[width=\sampleImageWidth]{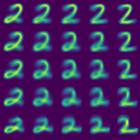} & \includegraphics[width=\sampleImageWidth]{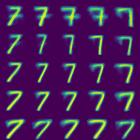} & \includegraphics[width=\sampleImageWidth]{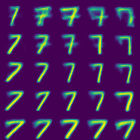} \\ 
    \bottomrule
    \end{tabular}
    \caption{Sample images for \vsud\ }
    \label{tab:sudoku_samples}
\end{table}
\pagebreak
\section{Ablation Numbers}
We include numbers for modified versions of \NSN{} in Table~\ref{tab:detailed_ablation} for the \vsud\ task. Note that for all modified versions we get zero task completion rate and no better than chance classification accuracy indicating that both the reward gradient and subsampling heuristic are necessary to solve the cold-start issue. REINFORCE with maxent regularization adds a maximum entropy regularization term $-\tau \pi_\theta(\textbf{a}|s_0) \text{log} \pi_\theta(\textbf{a}|s_0)$ to the $O_{RL}$ objective in Eq~\ref{eq:RL} to encourage exploration. We try $\tau \in \{1, 10, 100, 1000\}$. UREX \cite{Nachum2017} also uses a hyperparameter $\tau$ to characterize the reward-scaled probability distribution. We try $\tau\in\{0.005, 0.01, 0.1, 1\}$, values in the neighborhood of the optimal value $\tau=0.1$ recommended in the official implementation. 


\begin{table}[h!]
    \centering
    \begin{tabular}{cccc}
    \toprule
        \multirow{2}*{\textbf{Ablation}} & \multirow{2}*{\textbf{Hyperparameter}} & \multicolumn{2}{c}{\textbf{Sudoku}}   \\ 
         \cmidrule(lr){3-4} 

         & & CA & TCR \\
         \midrule
         \multirow{5}*{\NSN{}} & $\eta = 1$ & 11.77 & 0.0 \\
        & $\eta=\nicefrac{1}{2}$ & 20.92 & 0.0 \\
        & $\eta=\nicefrac{1}{16}$ & 38.65 & 0.0 \\
        & $\eta=\nicefrac{1}{32}$ & 98.35 & 60.9 \\
        & $\eta=\nicefrac{1}{128}$ & 98.56 & 65.4 \\
         \midrule 
         \NSN{} without reward gradient & - & 58.77 & 0.0  \\ 
         \midrule 
         \multirow{4}*{REINFORCE with maxent regularization} & $\tau=1$ & 11.08 & 0.0 \\ 
         & $\tau=10$ & 10.03 & 0.0 \\ 
         & $\tau=100$ & 10.57 & 0.0 \\
         & $\tau=1000$ & 10.53 & 0.0 \\ 
         \midrule 
         \multirow{4}*{UREX} & $\tau=0.005$ & 10.53 & 0.0 \\ 
         & $\tau=0.01$ & 11.77 & 0.0 \\ 
         & $\tau=0.1$ & 10.53 & 0.0 \\
         & $\tau=1$ & 10.57 & 0.0 \\ 
         \bottomrule
    \end{tabular}
    \caption{TCR and CA values for \NSN{} with ablations}
    \label{tab:detailed_ablation}
\end{table} 
\end{appendices}

\end{document}